%% file: main.tex
\documentclass[runningheads]{llncs}

\input{preamble}

\input{defs}

\begin{document}

\maketitle
\input{sec/0_abstract}    
\input{sec/1_intro}

\input{sec/2_Related_Work}
\input{sec/3_Method}
\input{sec/4_Results}

\input{sec/5_conclusion}

% \paragraph{Acknowledgements} T. Birdal acknowledges support from the Engineering and Physical Sciences Research Council [grant EP/X011364/1].


%\clearpage  % TODO REVIEW/FINAL: This \clearpage needs to be removed from both review and camera-ready versions.


\bibliographystyle{splncs04}
\bibliography{main}

\end{document}

%% file: preamble.tex
\usepackage{eccv}

\usepackage{eccvabbrv}
\usepackage{graphicx}
\usepackage{tikz}
\usepackage{multirow}
\usepackage{comment}
\usepackage{mathtools}

\usepackage{amsthm}
\usepackage{wrapfig}
\usepackage{enumitem}
\usepackage{booktabs}
\usepackage{bm}
\usepackage{color}
\usepackage{fancyvrb}
\usepackage{gensymb}
\usepackage{algorithm,algpseudocode}
\algrenewcommand\algorithmicrequire{\textbf{Input:}}
\algrenewcommand\algorithmicensure{\textbf{Output:}}
\usepackage{stackengine}
\stackMath

\usepackage[accsupp]{axessibility}  % Improves PDF readability for those with disabilities.
\usepackage{times}
\usepackage{epsfig}
\usepackage{graphicx}
\usepackage{amsmath}
\usepackage{amssymb}

\usepackage[utf8]{inputenc}
\usepackage[T1]{fontenc}
\usepackage{lmodern}
\usepackage{graphicx}
\usepackage{caption}
\usepackage{amssymb}% http://ctan.org/pkg/amssymb
\usepackage{pifont}% http://ctan.org/pkg/pifont
\newcommand{\cmark}{\ding{51}}%
\newcommand{\xmark}{\ding{55}}%
\usepackage{subcaption} % loads the caption package
\usepackage{float}
\usepackage{makecell}

\usepackage{dblfloatfix}

\usepackage{hyperref}
\usepackage[capitalize]{cleveref}
\usepackage{orcidlink}

\definecolor{ben}{rgb}{0.9,0.,0.5}

\title{SABER-6D: Shape Representation Based Implicit Object Pose Estimation} 
\author{Shishir Reddy Vutukur \inst{1,2} \orcidlink{0000-0002-4406-8491}  \and
Mengkejiergeli Ba\inst{3} \orcidlink{0009-0008-7905-9609} \and 
Benjamin Busam\inst{1,4}  \orcidlink{0000-0002-0620-5774}  
\and  \newline  Matthias Kayser\inst{3} \orcidlink{0009-0000-5228-3397} \and 
Gurprit Singh\inst{4} \orcidlink{0000-0003-0970-5835}
}
\authorrunning{Vutukur et al.}

\institute{$^1$Technical University of Munich  \quad $^2$Siemens Technology \quad $^3$Robert Bosch GmbH 
 \newline $^4$Munich Center for Machine Learning (MCML)  
 \quad $^4$Max-Planck Institute(MPI) }

%% file: defs.tex
\renewcommand{\vec}[1]{\mathrm{vec}{(#1)}}

\DeclarePairedDelimiterX{\infdivx}[2]{(}{)}{%
  #1\;\delimsize\|\;#2%
}

\crefname{equation}{eq.}{eq.}
\Crefname{equation}{Eq.}{Eq.}
\crefname{theorem}{thm.}{thms.}
\Crefname{Theorem}{Thm.}{Thms.}
\crefname{conjecture}{conj.}{conjs.}
\Crefname{Conjecture}{Conj.}{Conjs.}
\crefname{proposition}{prop.}{props.}
\Crefname{proposition}{Prop.}{Props.}
\crefname{definition}{dfn.}{dfn.}
\Crefname{definition}{Dfn.}{Dfn.}
\crefname{remark}{remark}{remark}
\Crefname{Remark}{Remark}{Remark}
\Crefname{algorithm}{Alg.}{Alg.}

\crefname{section}{Sec.}{Secs.}
\Crefname{section}{Sec.}{Secs.}
\crefname{equation}{Eq.}{Eqs.}
\Crefname{equation}{Eq.}{Eqs.}
\crefname{figure}{Fig.}{Figs.}
\Crefname{figure}{Fig.}{Figs.}
\crefname{table}{Tab.}{Tabs.}
\Crefname{table}{Tab.}{Tabs.}
\crefname{thm}{Thm.}{Thms.}
\Crefname{thm}{Thm.}{Thms.}
\crefname{conj}{Conj.}{Conjs.}
\Crefname{conj}{Conj.}{Conjs.}
\crefname{dfn}{Dfn.}{Dfns.}
\crefname{dfn}{Dfn.}{Dfns.}
\crefname{remark}{remark}{remarks}
\Crefname{Remark}{Remark}{Remarks}
\crefname{prop}{Prop.}{Prop.}
\Crefname{prop}{Prop.}{Prop.}
\Crefname{algorithm}{Alg.}{Alg.}
\crefname{appendix}{App.}{apps.}
\Crefname{appendix}{App.}{Apps.}
\crefname{appsec}{appendix}{appendices}
\Crefname{appsec}{Appendix}{Appendices}

\renewcommand{\paragraph}[1]{{\vspace{0.6mm}\noindent \bf #1}.}

%% file: sec/0_abstract.tex
%\section{Experiments}

% \end{experimentss}

\begin{abstract}
In this paper, we propose a novel encoder-decoder architecture, named SABER, to learn the 6D pose of the object in the embedding space by learning shape representation at a given pose. This model enables us to learn pose by performing shape representation at a target pose from RGB image input. We perform shape representation as an auxiliary task which helps us in learning rotations space for an object based on 2D images. An image encoder predicts the rotation in the embedding space and the DeepSDF based decoder learns to represent the object's shape at the given pose. As our approach is shape based, the pipeline is suitable for any type of object irrespective of the symmetry. Moreover, we need only a CAD model of the objects to train SABER. Our pipeline is synthetic data based and can also handle symmetric objects without symmetry labels and, thus, no additional labeled training data is needed. The experimental evaluation shows that our method achieves close to benchmark results for both symmetric objects and asymmetric objects on Occlusion-LineMOD, and T-LESS datasets.

% A translation estimation block is used to regress the translation in the object reference frame using the predicted rotation. We also propose a shape-based loss to handle symmetric objects indifferently by including shape representation in the pipeline. 
\end{abstract}

%% file: sec/1_intro.tex
\section{Introduction}
Information about the 6D pose of an object in the world is crucial for many applications such as robotics and augmented reality. 
In robotics, a precise pose is needed to locate an object for further operations such as grasping and automatic assembling.
Basically,
6D pose refers to estimating the translation and orientation of an object in 3D space from a viewpoint.
In our work, we aim to train a neural network \textit{SABER} to estimate the 6D pose using only the CAD model of the objects without any prior labels about the symmetry. 

Most research \cite{surfemb, hodan2020epos, sc6d, dpod, dpodv2, su2022zebrapose} in this field has a strong focus on bridging the gap between synthetic and real by employing training pipelines using data generated purely from CAD models using BlenderProc\cite{Denninger2023}. Our work also tries to address this issue and in addition, solve the problem of pose estimation without knowing the object's symmetry. Many approaches\cite{dpod, dpodv2, gdrn, pix2pose} assume the symmetry knowledge of an object beforehand and try to convert ambiguous poses for symmetric objects into unambiguous poses and regress correspondences or poses in the unambiguous space. This is impractical in some scenarios because estimating symmetry labels for a CAD model with higher degrees of symmetry is difficult. Moreover, objects like a coffee mug, which are conditionally symmetric, are much more difficult to deal with while estimating their pose. It is also difficult to annotate symmetry labels for such objects and for objects with complex symmetries as shown in the Symmetric Solids Dataset\cite{implicitpdf}. Thus, an approach that can handle any object irrespective of symmetry labels is desired.  
%
% \begin{figure}
%   \includegraphics[width=\linewidth]{images/TeaserUp.pdf}
%   \caption{Overview of SCOPE. We propose an Encoder-Decoder architecture to learn rotations in embedding space using just 3D CAD models. SCOPE is capable of shape completion and  6D pose estimation simultaneously using a learned generalized shape representation from point clouds.}
%   \label{fig:raygen}
% \end{figure}

Approaches like SurfEmb\cite{surfemb}, SC6D\cite{sc6d}, and Augmented Autoencoder (AAE)\cite{Sundermeyer2018} employ textured CAD models without assuming symmetry for the pose estimation task. We employ a similar setting by training the approach with CAD model based photorealistic data (PBR) generated using BlenderProc\cite{Denninger2023} without assuming symmetry labels are known.   
SurfEmb learns symmetry invariant per-point features to handle symmetry. During inference, they employ an intensive render and compare pipeline to estimate 6D pose from 2D-3D correspondences. They need the intensive inference step to handle symmetrical objects as they cannot get one-to-one correspondences for symmetric objects. SC6D assigns a embedding vector for each rotation in a sampled SO3 space and formulates a contrastive loss with a embedding vector from the CNN. They optimize both CNN and rotation embedding vectors during training. SurfEmb handles the symmetry ambiguity by learning symmetry invariant features while SC6D handles this by mapping viewpoint to a optimizable rotation embedding vector. SC6D performs slightly better than our approach as it solves a easier task to optimize rotation embeddings compared to the shape reconstruction problem that we employ. AAE learns to generate unaugmented images from an augmented version of the same image. This enables them to learn the pose in embedding space, but they do not make explicit use of shape information from the CAD model. We employ a similar pipeline to learn the pose in the embedding space. We generate the shape of the object from a 2D image at the same pose as the input image. This enables us to learn rotations in the embedding space. Concretely, our pipeline takes a single 2D image as input and predicts the shape of the object at the same rotation as the 2D image. To reconstruct the shape of the object at the same pose as the input image, the network has to learn rotations in the embedding space. This view-based shape prediction enables us to map ambiguous poses to a single embedding vector as both the shape and the image do not change with different symmetric configurations. 

Our work starts with the exploration of DeepSDF\cite{Park2019}, which is a method to learn the 3D representation of objects in a canonical orientation. DeepSDF learns to represent the shape of a 3D object by making the network learn the Signed Distance Function (SDF) of an object.
DeepSDF learns to represent objects in canonical space and performs shape completion in canonical space from partial point clouds. The shape completion and representation do not generalize to rotated shapes. They only operate in one canonical orientation. In our approach, we employ the DeepSDF decoder to learn object shapes at different rotations and learn to represent and reconstruct shapes at different rotations in SO3 space. During training, our network learns to represent rotations in embedding space. However, we observe that we cannot learn to predict the shape from the 2D image by simultaneously training a CNN encoder and the DeepSDF decoder. Thus, we employ a two-stage pipeline for training which involves initially training the DeepSDF decoder in an Autodecoder manner where embedding vectors for rotations are optimized along with the decoder itself. This pre-training helps in conditioning the decoder to learn to represent the shape at different rotations. Then, we train the encoder and decoder together that learns to represent the shape at the desired rotation based on the 2D image input. Some approaches \cite{welsa, d1,d2, centersnap, FSD} deploy DeepSDF as part of the pose estimation training pipeline, but they employ DeepSDF in canonical orientation and for learning a category of shapes. Contrarily, we employ DeepSDF to represent many rotated versions of the same instance object and to create a symmetry invariant pipeline.   

Concretely, we obtain the 2D image and pass it through a CNN to get a rotation embedding feature. The rotation embedding feature along with a 3D point is passed through a DeepSDF decoder to predict the SDF value of the point. Predicting the SDF value from a 2D image indicates that the network represents the shape at the pose conditioned on the input 2D image. Our pipeline enables the prediction of the rotated shape of an object from 2D image input. We make our implementation publicly available \href{https://github.com/shishirreddy/Saber6D}{here}.

The contributions of our work are summarized as follows:
\begin{itemize}
	\item We propose a novel approach, SABER, which implicitly estimates the 6D pose of the object in embedding space by conditioning shape prediction on the input image.
	\item Our network can handle CAD models without symmetry labels by employing a shape representation based approach. 
	% \item Additionally, in the case of known symmetry for continuous symmetric objects, we propose a novel loss function for handling ambiguous poses during training instead of handling them at the data preprocessing phase.
\end{itemize}

%
% \monote{
% What to write here:
% \begin{itemize}
% 	\item What's the paper about? What problem/context is adressed?
% 	\item Where can such an approach be used in practice? Why is it important to solve this problem?
% 	\item What problem is solved (only CAD model is needed for training) => why is this intresting?
% 	\item What is new in our paper? (only CAD model, learn a embedding code for the objects based on this. We can also reconstruct the objected based on a partial pointcloud => DeepSDF)
% 	\item Briefly state what others are doing or what are the current state-of-the-art.
% \end{itemize}
% }

%% file: sec/2_Related_Work.tex
\section{Related work}

\subsection{Pose estimation}
% Pose estimation approaches can be categorized as RGB based \cite{Sundermeyer2018, Zakharov2019, Park2019a, Labbe2020}, RGB-D based \cite{Xu2019, Chen2020, Wang2019} or depth-based \cite{Gao2020, Vidal2018, Hinterstoisser2016}. 

Many deep learning methods \cite{Zakharov2019, Park2019a, Chen2020, Peng2019, Xu2019} estimate the pose as a regression problem directly relying on the labeled pose. Considering the high cost of annotating labels, NOL \cite{Park2020} trains the networks using both real images and synthetic data. NOL applies a network to generate synthetic images covering various unseen poses from a few cluttered images and texture-less 3D models of objects. Real data based approaches, NeRF-Pose~\cite{nerfpose} and NeRF-Feat~\cite{nerfeat} employ NeRF to train pose estimation pipeline in the absence of a CAD model. Welsa-6D ~\cite{welsa} employs a feature based point cloud registration approach to label weakly labeled samples using few labeled data which also does not assume the symmetry label is known. TexPose~\cite{chen2023texpose} leverages real images for photometrically accurate textures while geometry is learned from synthetic renderings.  

 In AAE, different views of objects are rendered as training data. An augmented auto-encoder structure is utilized to reconstruct the object image and the reconstruction loss serves as the supervision signal to train the network. Our network also learns the pose implicitly which is accomplished by 3D shape prediction. Pix2Pose\cite{Peng2019}, DPOD\cite{ Zakharov2019} estimate the pose by directly learning 2D-3D correspondences and the pose is estimated in a refinement step, i.e., by PnP and RANSAC algorithms. In \cite{Park2019a}, pose estimation is treated as a pixel-wise 3D coordinates regression problem. For each pixel, the 2D-3D correspondence is obtained by modeling an object in a colored space from an RGB image and texture-less 3D model. CosyPose \cite{Labbe2020} starts with a single-view 6D pose estimation to select object candidates from multiple images of a scene to refine the estimated pose.

% RGB-D based methods use both RGB image and depth image as training data. W-PoseNet \cite{Xu2019} first generates a point cloud from RGB-D data, which are used to extract pixel-wise features. The outputs are a query of pixel-pair values, which consist of the orientation and confidence, and the best pose prediction is decided according to the confidence value. In the G2L-Net \cite{Chen2020}, the point cloud is given to a PointNet-based module \cite{Qi2017} to obtain a 3D segmentation and a translation residual prediction. Next, the point-wise embedding vector features are extracted, which go through two point-based decoders for the prediction of rotation and a rotation residual independently. Similarly, DenseFusion\cite{Wang2019} also extracts embedding vectors features to estimate the pose. The difference is that DenseFusion works on pixel-level features while G2L-Net deals with point features from 3D geometric information. 

% Depth-based methods do not rely on color information. The classical point-pair-feature (PPF) algorithm \cite{Drost2010} is a broadly used method to estimate 6D pose by global modeling and local matching. \cite{Hinterstoisser2016,Vidal2018} have proposed some improvements based on PPF. \cite{Hinterstoisser2016} modify the sampling and voting scheme in PPF to cope with the sensor noise and cluttered background issue. In \cite{Gao2020}, a semantic segmentation method is first applied on a depth image to generate an object point cloud. Then, the object point cloud is sent to two networks to estimate the rotation and translation separately. 

A critical issue in pose estimation is visual ambiguity~\cite{manhardt2019explaining}. Given a symmetric object, it remains the identical pose with reference to the camera when rotated about a symmetric axis, which leads to rotation ambiguity.  SurfEmb, NeRF-Feat, SC6D, Augmented AutoEncoder, and EPOS\cite{hodan2020epos}, MatchU\cite{matchu}, DiffusionNocs\cite{diffusionnocs} proposed approaches that can handle symmetries by design without assuming symmetry labels. We also propose an approach in a similar direction where we do not assume symmetry labels are known. We compare our approach to RGB based methods, SurfEmb, SC6D, AAE, EPOS, without requiring depth.

Aforementioned methods focus solely on an instance level pose estimation. Category level pose estimation approaches typically employ a shape estimation/deformation module to predict the shape of unseen instances. Shapo \cite{d2}, iCaps, \cite{deng2022icaps}, SGPA \cite{chen2021sgpa}, Zero123-6D\cite{di2024zero123}, FSD \cite{FSD}, CenterSnap\cite{centersnap} employ shape prediction modules in canonical orientation while our approach employs shape prediction at the same orientation as the object in the image. 

There is a line of research where rotation distribution is estimated to understand the symmetries of objects without prior labels. Implicit-PDF\cite{implicitpdf}, SpyroPose\cite{haugaard2023spyropose}, HyperPosePdf\cite{hofer2023hyperposepdf} employ approaches to learn rotation distribution for challenging symmetric objects without assuming symmetry labels. Our approach focuses on 6D pose estimation, but it can potentially be extended to estimate rotation distribution similar to these approaches.    
% \subsection{Shape representation}

% Conventionally, there are three ways to represent a 3D object’s shape: point cloud, voxels, and meshes. As described in \cite{Xu2019a}, all three representations have some limitations to achieve a high-quality object’s shape. One approach that addresses these limitations is to model the 3D surface of an object using a continuous function. Both \cite{Park2019} and \cite{Xu2019a} train the networks to model a signed distance function (SDF), which estimates the signed distance between a space point and an object‘s surface. \cite{Park2019} applies an auto-decoder structure to form the network. Object’s shape information is embedded as a latent vector and fed into the network as input. \cite{Xu2019a} trains the network from both local features and global images based on an auto-encoder structure. \cite{Mescheder2019} further simplifies the problem as a binary classification. That is to train a deep neural network to determine if a space point is occupied by the object’s surface by assigning a binary value.
%\monote{
%	What to write here:
%	\begin{itemize}
%		\item Describe other approachs. I would structure it topic-wise: Shape Completion/3D representation and 6D Object Pose Estimation
%		\item here we should also do a comparison to them, what have in common or what we have did in other new and better way.
%		\end{itemize}}

%% file: sec/3_Method.tex
%  macros definitions
\newcommand{\xni}[3]{#1_{#2_{#3}}}
\newcommand{\xki}[3]{#1^#2_{#3}}
\newcommand{\xn}[2]{#1_{#2}}
\newcommand{\xnik}[4]{#1^{#4}_{#2_{#3}}}
\newcommand{\latT}[1]{T^{#1}}
\newcommand{\latO}[1]{O^{#1}}

\section{Approach}
\label{sec:approach}
Our approach, SABER, employs a two-stage encoder-decoder network with a rotation estimation block followed by a translation predictor for the 6D pose estimation.
%Our approach comprises an encoder-decoder based architecture to estimate rotation and a Translation Predictor network to estimate translation offset.
The network learns the shape representation of objects in embedding space at different orientations, which are used to predict the rotation. 
The encoder-decoder pipeline learns rotations in embedding space and the signed distance function (SDF) of the object at different orientations.
Learning SDF values enables us to perform shape representation and employ a shape-dependent loss function for pose estimation, which helps to treat symmetric objects indifferently. 

Unlike other approaches, where the encoder and decoder are trained together, we train the decoder first and then the encoder together with the already pre-trained decoder. We choose this paradigm to make the decoder initially learn rotation space which can then generalize well and adapt well to unseen rotations during the second stage.
The decoder initially learns to represent the object at some randomly sampled rotations and learns rotations in embedding space. DeepSDF optimizes object related embedding vectors during training while we optimize rotation related embedding vectors during training the decoder. However, these optimized embedding vectors are only used for training in the first stage and are discarded for the second stage.
% \begin{figure}
%   \includegraphics[width=0.5\linewidth]{images/raygenflip.png}
%   \caption{Partial Point Cloud Architecture}
%   \label{fig:raygen}
% \end{figure}

% \begin{figure}
%      \centering
%      \begin{subfigure}[b]{0.3\textwidth}
%          \centering
%          \includegraphics[width=\textwidth]{images/pc2.png}
%         %  \caption{$y=x$}
%          \label{fig:y equals x}
%      \end{subfigure}
%      \hfill
%      \begin{subfigure}[b]{0.3\textwidth}
%          \centering
%          \includegraphics[width=\textwidth]{images/mesh.png}
%         %  \caption{$y=3sinx$}
%          \label{fig:three sin x}
%      \end{subfigure}
%      \hfill
%      \begin{subfigure}[b]{0.3\textwidth}
%          \centering
%          \includegraphics[width=\textwidth]{images/rays2.png}
%         %  \caption{$y=5/x$}
%          \label{fig:five over x}
%      \end{subfigure}
%     \caption{partial point cloud generation}
%     \vspace{-1.5em}
%     \label{fig:pcgen}
% \end{figure}
%

\subsection{6D Pose Estimation}
The encoder takes an image as input and predicts the rotation in embedding space. The decoder decodes the shape from the rotation embedding at the same pose as the input image. After training, we discard the decoder and create a lookup table with rotation embedding vectors generated from the encoder using different training images with corresponding rotation matrices. During Inference, the embedding vector is predicted from the image encoder, and the rotation matrix corresponding to the closest embedding vector in the lookup table is considered the estimated rotation. Similar to GDR-Net and SC6D, we also employ scale invariant translation estimation (SITE) coordinates to estimate translation from a 2D image using a CNN, Translation predictor.  In effect, we estimate the rotation from the encoder and the translation from the translation predictor to get the final 6D pose.

% We can reconstruct the shape at any given pose from the image using the predicted rotation embedding and the decoder. By uniformly sampling points in a unit cube and estimating their corresponding SDF values using the decoder and the embeddings, a complete mesh can be reconstructed using the Marching Cubes algorithm \cite{Lorensen1987}. 

\begin{figure*}[h]
	\centering
	\includegraphics[width=\linewidth]{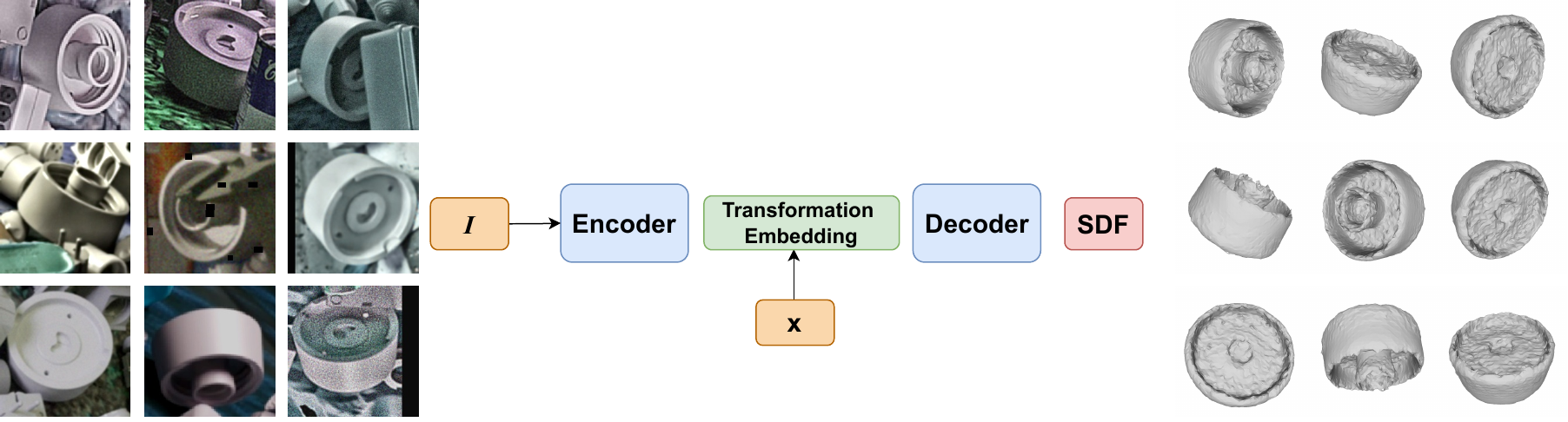}
	\caption{Encoder-decoder architecture. A 2D image, $I$ is input to the image encoder to predict transformation embedding. A 3D point, $x$, is concatenated to the transformation embedding and input to the decoder to predict the corresponding SDF value. The input images, the corresponding learned shape representations are visualized by generating meshes using marching cubes. Meshes can be generated by estimating SDF values for 3D points in a grid. This pipeline is employed for estimating rotation.}
	\label{fig:architecture}
\end{figure*}
\subsection{Architecture}   
Our encoder-decoder architecture is depicted in Figure \ref{fig:architecture}. The architecture comprises an image based encoder and a multilayer perceptron as the decoder. The translation predictor uses another CNN network to regress the translation. These blocks are explained in the following sections. We employ an Allocentric representation similar to GDR-Net and SC-6D as allocentric rotation is defined based on object rotation and not based on viewpoint.
% This is achieved by training the decoder to learn rotations and shape of the object in embedding space irrespective of the type of symmetry of the object followed by training the encoder during which we use different loss functions for symmetric objects to handle pose ambiguities in estimating correspondences.

% It uses a feed-forward network with an embedding vector and a 3D point as input to predict a Signed Distance Function (SDF) value of the point. The network learns to represent shapes with an embedding vector assigned to each shape by learning SDF value for 3D points in space for each object.

\subsubsection{Decoder}
%  We employ a decoder to learn shape representation as well as orientations for different objects. We extend the architecture introduced by the shape representation network DeepSDF\cite{Park_2019}. However, DeepSDF only learns the shape representation in canonical orientation, and hence shape completion does not work for different orientations. We introduce another embedding space for orientations to address this issue. To enable pose estimation, we add correspondences into the pipeline. These changes allow our network to do shape completion as well as pose estimation simultaneously. 
 
% %  The network uses 7 fully connected layers with Relu activation. The input vector is concatenated to intermediate output after 4th layer and the input 3D point is also concatenated to every intermediate output for better feature learning.  
% %   This setup enables us to simultaneously learn correspondences and shape representation of the object at different orientations.

% Two different embedding spaces are optimized during training, one corresponding to the shape of the object and another corresponding to the object's orientation. In addition to predicting SDF values, we also predict the corresponding coordinate of the transformed input point in canonical orientation. The decoder is trained in an Auto-decoder manner \cite{Park2019} without an encoder where embedding vectors are optimized along with the network's weights during the training.
%We employ a decoder to learn shape representation as well as orientations for different objects.
As mentioned before, our decoder is adopted from the DeepSDF approach. DeepSDF initially samples random embedding vectors for representing each object in the category that are optimized along with the network. 
Analogously, we use different embeddings for representing the object at different rotations which are optimized along with the decoder similar to DeepSDF. 
Our decoder is trained as an auto-decoder, without an encoder, similar to DeepSDF where embedding vectors are optimized along with the weights during training.
 
Specifically, we sample $K$ embedding vectors corresponding to $K$ rotations with a dimension of $1000$ from $\mathcal{N}(0,1/1000)$, which we will call transformation embeddings. Although, the embedding vectors are sampled randomly, the embeddings are optimized during the decoder training. We randomly sample $K$ quaternions to learn objects at $K$ orientations. 
 
To learn the SDF of an object, we sample 3D points in the space around and inside the object and estimate their SDF values from the object mesh at canonical orientation to generate ground truth training samples. We can generate training samples for different rotations by simply rotating the 3D points. Rotated 3D points with the same SDF values represent rotated shape. We refer to the 3D points as $X$ and the corresponding SDF values as $S$ in the canonical orientation. We refer to a single 3D point as $x$ and a vector of 3D points as $X$ and we follow this convention for other variables as well. 

The training samples constitute a transformation embedding vector, $\latT{}_{k}$, corresponding to $k$-th rotation, a 3D point, $x_{k}$, rotated with $k$-th rotation and its corresponding SDF value, $s$. Note that the SDF value does not change with object rotation as the rotated 3D points with the same sdf values represent the rotated object.  
% Let $x$ be a 3D point in the canonical space of the object.
Each transformation embedding vector, $\latT{}_{k}$, is assigned to a rotation, $R_{k}$. The 3D point sample corresponding to the object rotated by $k$-th rotation, $R_{k}$, is concatenated with corresponding transformation embedding $\latT{}_{k}$ to create the input vector. $\phi$ is a one-to-one mapping function between embedding vectors and corresponding rotation matrices. The network, $D_{\theta}$, takes the input vector and predicts the SDF value, $\xnik{s}{}{}{}$.
\begin{equation}
	\begin{split}
		\xn{R}{k}&= \phi(\latT{}_{k}), \quad
		\xnik{x}{k}{}{}=\xn{R}{k} \cdot x\\
		% \xnik{C}{n}{}{k} &= (\xnik{x}{n}{}{k},\latO{}_{n},\latT{}_{k})
		\xnik{s}{}{}{} &= D_{\theta}(\xnik{x}{k}{}{},\latT{}_{k})
	\end{split}
\end{equation}

\subsubsection{Encoder} \label{encoder}
A ResNet16 \cite{resnet} encoder extracts features from an RGB image and predicts the transformation in the embedding space.   
% Input to the encoder is obtained by concatenating the one-hot vector to each 3D point. We also append the one-hot vector to the global feature and intermediate outputs after the second, third set abstraction layers. 
The transformation embedding is input into the decoder along with the 3D points to predict the corresponding point-wise SDF value.

%We choose to send partial point cloud through the encoder because the input to network during inference is similar to partial point cloud.
%However, we send full point cloud through the decoder to employ loss functions defined for the full shape of the object.
%This design helps to apply symmetry specific loss functions for symmetric objects during training the encoder. 
The image corresponding to $k^{th}$ rotation, $\xnik{{I}}{k}{}{}$, is input to the encoder to predict transformation embedding $\latT{}_{k}$ as follows:
\begin{equation}
	% \begin{split}
	% \xnik{{P}}{n}{}{k}=(\xn{R}{k}\cdot \xni{{I}}{k}{}), \quad
	\latT{}_{k} = E_{\theta}(\xnik{{I}}{k}{}{})
	% \end{split}
\end{equation}
% \begin{equation}
% \end{equation}
% To disentangle rotation and translation while estimating rotation, we train the encoder by sending the partial point cloud at same orientation with random translations to make the network invariant to translations.

% \subsubsection{Translation predictor}
% \begin{figure}[t]
%   \includegraphics[width=\linewidth]{images/transPredictor_FinalV3.pdf}
%   \caption{Translation predictor module.}
%   \vspace{-0.5em}
%   \label{fig:tranlationPredictor}
% \end{figure}
\subsubsection{Translation predictor}
This module estimates the translation in SITE coordinates as employed in GDR-Net and SC6D. SITE coordinates comprises a 2D offset between the center of the object and the center of the image crop and z-axis translation. We employ a CNN similar to the one employed in SC6D for predicting SITE coordinates from a 2D image. The 2D image corresponding to the $k^{th}$ rotation, $\xnik{{I}}{k}{}{}$, is input to the translation predictor, $M_{\theta}$  to predict SITE coordinates, $t_{k}$, as follows:
\begin{equation}
\xnik{t}{k}{}{} = M_{\theta}(\xnik{{I}}{k}{}{})
\end{equation}
% \subsection{Encoder} \label{encoder}

\subsection{Training Stages}
% \subsubsection{Decoder Loss functions} \label{decoderLoss}
We train our pipeline in two stages instead of directly training both decoder and encoder together. We train the decoder in the first stage and then we train both the encoder and decoder in the second stage.

We employ a two stage pipeline because we observed that when the encoder and decoder are trained together in a single stage, it falls into local minima and reconstructs sphere shape for all poses. The sphere shape is a local minima since we are rotating our object around the origin and the sphere will locally minimize all the rotated shapes in a batch and predict a sphere to satisfy them all. The task of shape estimation at various poses based on the input image is difficult and simultaneously training the encoder and decoder from scratch does not facilitate the shape learning. To avoid this, we sample some rotations randomly and train our decoder alone to represent and learn the rotated versions of our object. This makes it easier for the network to learn shapes directly without also optimizing the encoder. This pre-training of the decoder helps it to learn the shape and also understand the rotation space.  

\textbf{Stage 1:}
We train the DeepSDF decoder in an auto-decoder manner where rotation embedding codes are optimized along with the decoder network. After training the decoder, the decoder learns to represent shapes at different rotations. We discard the learned embedding vectors representing rotations after this stage as we train an encoder in the second stage to predict the embedding vectors.

The decoder takes a concatenated vector of 3D point and the embedding vector to predict the SDF value. The shape loss, $L_{s}$ is formulated to predict accurate SDF values. The ground truth SDF values, predicted SDF values are denoted as  $\xnik{\hat{S}}{}{}{}$ and  $\xnik{{S}}{}{}{}$ respectively. The decoder loss, $L_{D}$, used to train the decoder computed as follows: 
\[
L_{D}=\sum_{}||\xnik{S}{}{}{},\xnik{\hat{S}}{}{}{}||_{1}
% , \quad L_{c}=\sum_{n,k}||\xnik{Y}{n}{}{k},\xnik{\hat{Y}}{n}{}{k}||_{1}
\]
% \begin{equation}
% 	L_{D}= \alpha L_{s} + \beta L_{c}
% \label{decoderLoss}
% \end{equation}
% The total decoder loss, $L_{D}$, used to train the decoder is computed as follows:
% \begin{equation}
% 	L_{D}= L_{s} + L_{c}
% \end{equation}
% The Shape loss enables the network to learn shape representation, and learning to predict SDF values enables shape completion and also to better learn the transformation embedding space based on shape. The Correspondence loss enables the network to learn correspondences better, which are used for pose estimation. The one-hot vector of the object and the partial point cloud are sent through the encoder to predict the transformation embedding. The predicted transformation embedding, object embedding and the full point cloud are sent through the decoder to predict SDF, full point cloud at canonical orientation. We use the same predicted transformation embedding code of an object for each point in the full point cloud.

% \subsubsection{Encoder Loss functions} \label{encoderLoss}
\textbf{Stage 2} 
In this stage, we train the encoder and the pre-trained decoder together to represent shapes from rotations corresponding to the pose in the image. As the decoder is already pre-trained, the encoder finds it easy to predict shapes at various orientations. In this stage, the image encoder and decoder learn to map rotations from 2D image to shape representation through the transformation embedding.     
The image of the object corresponding to $k$-th rotation, $\xnik{I}{k}{}{}$, is input to encoder to predict the transformation embedding, $\latT{}_{k}$. The transformation embedding is concatenated with a 3D point, $\xnik{x}{k}{}{}$, and input to the decoder to predict corresponding per-point SDF values, $\xnik{s}{}{}{}$.
%  The encoder predicts transformation embeding as shown in eq(\ref{encoderLoss}). The transformation embedding, $\latT{}_{k}$, along with object embedding, $O_{n}$, is concatenated with each point in full point cloud, $\xnik{X}{n}{}{k}$, and input to decoder to predict SDF values, $\xnik{S}{n}{}{k}$, and 3D point correspondences, $\xnik{Y}{n}{}{k}$. 

We use the SDF loss, $L_{E}$ to train the encoder-decoder in this stage. SDF is not affected by symmetries since it is an inherent property of shape where symmetries are considered implicitly. 
%SDF does not change when a symmetric object is rotated about its symmetric axis since SDF is defined by appearance.
 \[
L_{E}=\sum_{}||\xnik{S}{}{}{},\xnik{\hat{S}}{}{}{}||_{1}
% , \quad L_{c}=\sum_{n,k}||\xnik{Y}{n}{}{k},\xnik{\hat{Y}}{n}{}{k}||_{1}
\]

%  Loss is calculated for decoder outputs and back-propagated through the decoder to the encoder to update the weights of the encoder
%
% We use different point sets as inputs to our decoder and encoder while training the encoder. We send partial point clouds into encoder and full point cloud at the same orientation into the decoder.

% \subsubsection{Translation Predictor Loss functions}
\textbf{Translation predictor loss functions:}
The translation predictor takes the 2D image corresponding to the $k^{th}$ rotation, $\xnik{{I}}{k}{}{}$, as input and predicts the SITE coordinates, $\xnik{t}{k}{}{}$. The translation predictor is trained with the loss function using the ground truth translation labels, $\xnik{\hat{t}}{k}{}{}$ as follows:
\begin{equation}
	L_{T}= \sum_{k} ||(\xnik{t}{k}{}{},\xnik{\hat{t}}{k}{}{})||_{1}
\end{equation}

%% file: sec/4_Results.tex
\section{Experiments}

% LineMOD \cite{Hinterstoisser2013}
We evaluated SABER on three datasets for 6D pose estimation: LineMOD \cite{Hinterstoisser2011}, Occlusion-LineMOD \cite{Brachmann2014}, T-LESS \cite{hodan2017tless}. We employ the standard AR metric from BOP challenge \cite{bopchallenge} for T-Less and Occlusion-LineMOD while we employ ADD score for LineMOD. For these datasets, we have trained SABER using only the PBR images generated from CAD models provided in the BOP challenge. We did not use the symmetry labels provided for objects in these datasets. We use the object detections from CosyPose\cite{Labbe2020} to have the same base data for a fair comparison.  Our pipeline achieves close to the state-of-the-art approach, SC6D, on LineMOD, Occlusion-LineMOD and T-LESS.  

\subsection{Training Protocol}

We train the decoder for $500$ epochs from scratch with a batch of $25$. Different learning rates are used to train the embedding vectors, $0.0005$, and the weights of the network, $0.001$, because we want the embedding vectors to vary less as the training progresses. Then, we use the trained decoder to train the encoder using PBR data for $2400$ epochs with a batch-size of $25$. The learning rate is set to $0.001$ while training the encoder-decoder in Stage 2. We use the same configurations as the encoder for the translation predictor. Our approach takes 30ms during inference for a single frame on an Nvidia Titan X.

\subsection{Architecture Details}
We employ a seven-layer MLP with ReLu activations for our Decoder. We employ a ResNet16 as the image encoder. For the translation predictor, we employ an SC6D based U-Net network with a ResNet34 encoder and decoder with 2D conv layers and 3 MLP layers.

% \subsection{Metrics}
\subsection{Evaluation}
For inference, we generate a codebook mapping rotation matrices to rotation embedding vectors using images from PBR data. We only use an image encoder to generate embedding vectors and discard the decoder for inference. We generate a codebook containing 200000 embedding vectors with corresponding rotation matrices. During inference, we send an image through the encoder to estimate an embedding vector which is used to find the nearest neighbor from the codebook and extract the corresponding rotation matrix as the estimated rotation. We estimate translation using the translation predictor. We employ the ADD score for evaluating LineMOD and the AR score from BOP for evaluating LineMOD-Occlusion and T-Less. During inference, we use the available detection crops from the models in CosyPose \cite{Labbe2020} trained on the synthetic images. 
\begin{table}[b]
   \centering
	\begin{tabular}{l|c|c|c|c|c|c}
		 \textbf{} & \textbf{GDRN}\cite{gdrn} &\textbf{Dpod}\cite{dpod} &\textbf{DpodV2}\cite{dpodv2}  &\textbf{AAE}\cite{Sundermeyer2018}  &\textbf{SC6D}\cite{sc6d}&\textbf{Ours}\\ \hline
   	 % \textbf{} & \cite{gdrn}    &\cite{sc6d}&\\ \hline
   
           \textbf{Sym} & \textbf{\cmark} & \textbf{\cmark} & \textbf{\cmark} & \textbf{\xmark}  &\textbf{\xmark}&\textbf{\xmark}\\\hline
           % \textbf{Ref} & \xmark & \cmark & \cmark &\xmark&\xmark\\
           
% 		 \hline
% 		 Ape		& 80.2 &53.3 & 81.4&69.97\\
% 		 Can		& 90.1 &86.5 & 94.7&83.08\\
% 		 Cat		& 61.2 &73.4 & 55.2&56.66\\
% 		 Driller	& 94.8 &92.8 & 86.0&86.31\\
% 		 Duck		& 77.6 &62.8 & 79.7&64.53\\
% 		 Egg box	& 72.9 &95.3 &65.6 &87.64\\
% 		 Glue		& 77.5 &92.5 &52.1 &78.71\\
% 		 Puncher	& 92.9 &76.7 &95.5 &89.16\\
		 \hline
		 \hline
		 ADD    & 0.77 & 0.66 & \textbf{0.81}& 0.31   &\textbf{0.73} & 0.71\\
	\end{tabular}
	\caption{Results on LineMOD. Sym refers to employing symmetry priors for training. GDRN refers to GDR-Net approach.}	
	\label{table:linemod}
% 			\vspace{-1.0em}
\end{table}

\subsubsection{LineMOD}

LineMOD dataset contains $15$ objects with some texture-less objects. It is a standard dataset to test the approach containing less occlusions, but it is still challenging since there is a Sim-to-real gap when we train on PBR data and evaluate on real scenes. We observed that our approach can achieve closer to benchmark approaches as shown in Table \ref{table:linemod} using ADD score. SC6D performs slightly better than our approach, while GDR-Net and Dpodv2 perform even better in the presence of symmetry labels. The performance improvement over AAE shows that shape encoding is stronger than image encoding.

\begin{table}[b]
   \centering
	\begin{tabular}{l|c|c|c|c|c|c}
		 \textbf{} & \textbf{GDRN}\cite{gdrn} &\textbf{CP}\cite{Labbe2020} &\textbf{EPOS}\cite{hodan2020epos} &\textbf{ SEMB}\cite{surfemb}  &\textbf{SC6D}\cite{sc6d}&\textbf{Ours}\\ \hline
   	 % \textbf{} & \cite{gdrn} &\cite{Labbe2020} &\cite{hodan2020epos}  &\cite{surfemb}  &\cite{sc6d}&\\ \hline
   
           \textbf{Sym} & \textbf{\cmark} & \textbf{\cmark} & \textbf{\xmark} & \textbf{\xmark} &\textbf{\xmark}&\textbf{\xmark}\\\hline
           \textbf{Ref} & \xmark & \cmark &\xmark & \cmark &\xmark&\xmark\\
           
% 		 \hline
% 		 Ape		& 80.2 &53.3 & 81.4&69.97\\
% 		 Can		& 90.1 &86.5 & 94.7&83.08\\
% 		 Cat		& 61.2 &73.4 & 55.2&56.66\\
% 		 Driller	& 94.8 &92.8 & 86.0&86.31\\
% 		 Duck		& 77.6 &62.8 & 79.7&64.53\\
% 		 Egg box	& 72.9 &95.3 &65.6 &87.64\\
% 		 Glue		& 77.5 &92.5 &52.1 &78.71\\
% 		 Puncher	& 92.9 &76.7 &95.5 &89.16\\
		 \hline
		 \hline
		 AR    & \textbf{0.71} &0.63 & 0.44 &0.66  &\textbf{0.59}& 0.55\\
	\end{tabular}
	\caption{Results on Occlusion-LineMOD. Sym and Ref refer to symmetry priors and refinement respectively.  GDRN, CP, SEMB refer to GDR-Net, CosyPose and SurfEmb.}	
	\label{table:OCCLinemod}
% 			\vspace{-1.0em}
\end{table}
\subsubsection{Occlusion-LineMOD}

The Occlusion-LineMOD dataset consists of $8$ objects from LineMOD with more challenging test scene. A scene from the original linemod dataset with heavy occlusions is extracted to test the performance in the presence of occlusions. This dataset is challenging due to severe occlusions and it comprises texture-less objects and symmetric objects. The occlusions are efficiently handled by the PBR data generated using realistic occlusions and it reflects in the performance of our approach in the presence of occlusions. We observed that SABER is robust to occlusions because we formulate the loss to reconstruct the full shape even from the occluded data and thus the encoder learns to handle occlusions well. We achieved 0.55 AR score, comparable to other SOTA approaches, as shown in Table \ref{table:OCCLinemod}. SurfEmb reports the results with refinement and also takes 2.2s for inference while our approach takes 30ms and doesn't employ refinement. GDR-Net has the best performance in the BOP challenge, but they assume that symmetry labels are known. Our approach achieves close to benchmark results from SC6D without symmetry prior. 

\begin{table*}[h]
	\begin{tabular}{l|c|c|c|c|c|c|c|c|c|c}
	 % \textbf{\makecell{Method/\\Symmetry}}
  \textbf{Method}& \textbf{GDRN} & \textbf{CP}& \textbf{CP}& \textbf{AAE}&\textbf{EPOS}& \textbf{SEMB} &\textbf{SEMB} & \textbf{SC6D} &\textbf{SC6D}&\textbf{Ours} \\
  \hline
  \textbf{Sym}& \cmark&\cmark&\cmark&\xmark&\xmark&\xmark&\xmark& \xmark&\xmark&\xmark\\
  \hline
   \textbf{Ref} & \xmark &\xmark &\cmark& \xmark &\xmark &\xmark& \cmark&\xmark&\cmark&\xmark \\
	 % \hline
	 % \hline
	 % RGB & \cmark & \cmark &\xmark &\xmark & \xmark \\
	 % ICP & \xmark & \cmark &\cmark &\cmark & \cmark \\
	 % Denoise & \xmark & \xmark&\xmark& \xmark & \cmark \\
	 \hline
	 
% 	 CS &33.68  &82.06 &63.27  &\textbf{87.22}& \textbf{86.73}\\
% 	 AS &52.96	&85.59&71   &76.91& 83.32 \\
% 	 DS &38.80  & 85.13&67.87 &80.70&84.83 \\
	 % Continuous &33  &82 &63  &\textbf{86}& \textbf{86}\\
	 % Asymmetric &52	&85&71   &76& 83 \\
	 % Discrete &38  & 85&67 &81&84 \\	 

	 \hline
	 Accuracy&\textbf{0.82} &0.52&0.64&0.30 &0.47 &0.62&0.74& \textbf{0.73}& 0.74 & 0.68\\
	\end{tabular}
	\caption{T-LESS Results. Sym and Ref
refers to employing symmetry priors and pose
refinement respectively.} 
	\label{TlessResults}
\end{table*}
\subsubsection{T-LESS}

The T-LESS dataset is an industrial dataset comprising $30$ texture-less objects with $11$ continuous symmetric objects, $16$ discrete symmetric objects, and $3$ asymmetric objects. This dataset involves $20$ different scenes with various levels of occlusion and clutter. We achieved an object recall accuracy of $0.67$. We compare our method with other SOTA approaches in the Table \ref{TlessResults}. We compare with other approaches which don't assume that symmetry is known beforehand. We observe that our approach achieves close to benchmark results by SC6D. We perform better than CosyPose which is also a correspondence-free method that assumes the symmetry labels are known. We also perform better than SurfEmb which also doesn't assume symmetry labels, and the inference time for SurfEmb is 2.2s compared to our approach which takes 30ms which is the same as SC6D. Our approach performs better than AAE which also works on similar principles as us by learning pose in latent space, but the performance of our shape based approach is better than the viewpoint image prediction method in AAE.

\subsection{Performance gap from SC6D}
While our approach achieves closer to the benchmark approach, SC6D, there is still some performance gap. This is because our approach tries to do a harder task compared to SC6D. Our approach tries to learn to reconstruct the object at different orientations which is a harder task compared to optimizing a rotation embedding for a viewpoint in SC6D. Although we didn't demonstrate the generalization capabilities of the approach, we believe our approach can be extended to a generalizable approach unlike SurfEmb and SC6D which are limited by the requirement of an MLP and rotation embeddings for every object they train on.       

\subsection{Ablation Studies} \label{ablation}

 We perform ablation on 4 objects(4,22,23,30) in T-LESS consisting 2 continuous symmetric, a discrete symmetric and an asymmetric object.

We perform an ablation to justify the choice of our two-stage pipeline. Approach trained with a single stage achieves only $5\%$ compared to our two-stage approach which achieves a $74\%$  as shown in Table \ref{table:ablation}.

We perform ablation for our choice of shape loss, Signed Distance Function loss over other shape loss, 
 Chamfer Distance. We regress correspondences instead of signed distance function and formulate chamfer distance as loss between estimated points and predicted points. We learn the rotations in latent space and follow the same inference pipeline to estimate rotation from latent space. Chamfer distance is also a shape based loss function, but the performance gap of  $59\%$ indicates that the Signed Distance Function loss is a stronger shape based loss compared to Chamfer distance in our scenario. 
 
We perform an ablation to justify our shape-based pipeline instead of a correspondence regression pipeline in the absence of a symmetry prior. We regress 3D correspondences instead of regressing the SDF value to show that our approach handles symmetric objects using a shape-based approach without symmetry prior. We regress correspondences and learn the rotations in latent space and follow the same inference pipeline to estimate rotation from latent space. We observe that the performance degrades by $24\%$ as the regression cannot handle symmetric shapes without prior information about symmetries.
\begin{table}[h]
\centering
	\begin{tabular}{l|c|c|c|c}
		 % \textbf{Method} & \textbf{GDRN} & \textbf{ SurfEmb} & \textbf{CosyPose} &\textbf{SC6D}&\textbf{Ours}\\ \hline
           \textbf{Exp} & \textbf{Reg} & \textbf{Cha} & \textbf{Ours-S} &\textbf{Ours-D}\\\hline
           % \textbf{Stage} & \textbf{1} & \textbf{1} & \textbf{1} &\textbf{2}\\
           
% 		 \hline
% 		 Ape		& 80.2 &53.3 & 81.4&69.97\\
% 		 Can		& 90.1 &86.5 & 94.7&83.08\\
% 		 Cat		& 61.2 &73.4 & 55.2&56.66\\
% 		 Driller	& 94.8 &92.8 & 86.0&86.31\\
% 		 Duck		& 77.6 &62.8 & 79.7&64.53\\
% 		 Egg box	& 72.9 &95.3 &65.6 &87.64\\
% 		 Glue		& 77.5 &92.5 &52.1 &78.71\\
% 		 Puncher	& 92.9 &76.7 &95.5 &89.16\\
		 \hline
		 \hline
		 AR     &0.5 &0.15 & 0.05& 0.74\\
	\end{tabular}
	\caption{Ablation Results on 4 objects in T-Less dataset. Reg refers to the approach trained using regression loss by predicting correspondences in canonical orientation and formulating a coordinate regression loss. Similarly, Cha refers to the approach trained using chamfer loss by predicting 3D points instead of SDF. Ours-S refers to our approach trained with a Single stage instead of employing a two-stage pipeline(Ours-D). }	
	\label{table:ablation}
% 			\vspace{-1.0em}
\end{table}

%% file: sec/5_conclusion.tex
\section{Conclusion \& Summary}
In this paper, we proposed a novel method, SABER, to estimate pose in implicit space by learning to represent shape from image input. We employ a DeepSDF based shape representation network to learn to represent an object at various orientations which enables us to learn poses in embedding space. Rotation embedding space is learned by employing shape and image which don't change with symmetry. Employing an implicit rotation estimation model enables us to handle symmetrical objects by design without requiring symmetry labels. We achieve close to state-of-the-art on Occlusion-LineMOD and T-Less datasets.  

% 1. Train two approaches-> Regression-based if you know symmetry
% 2. SDF-based if you don't know symmetry
% 3. Ablations: SDF vs Chamfer Distance
% 4. Regression vs special loss functions based on symmetry
% 5. Translation predictor ablation

% Ablation: 
% Two-stage vs 1 stage
% translation predictor ablation
% Tsdf vs sdf
% SDF vs Chamfer Distance

% \newpage